\documentclass{article}

\usepackage[final]{corl_2023} 

\usepackage{todonotes}
\usepackage{cleveref}
\usepackage{booktabs}
\usepackage{multirow}
\usepackage{array}
\usepackage{wrapfig}

\title{A multi-modal table tennis robot system}

\author{
  Andreas Ziegler\thanks{Equal contribution \qquad Correspondence to \{andreas.ziegler, \; thomas.gossard\}@uni-tuebingen.de\\This research is funded by Sony AI.}, Thomas Gossard$^{*}$, Karl Vetter, Jonas Tebbe, Andreas Zell\\
  Coginitive Systems Group\\
  Dept. Informatics\\
  University of Tübingen\\
}

\begin{document}
\maketitle


\begin{abstract}
    In recent years, robotic table tennis has become a popular research challenge for perception and robot control.
    Here, we present an improved table tennis robot system with high accuracy vision detection and fast robot reaction.
    Based on previous work, our system contains a KUKA robot arm with 6 DOF, with four frame-based cameras and two additional event-based cameras.
    We developed a novel calibration approach to calibrate this multimodal perception system.
    For table tennis, spin estimation is crucial.
    Therefore, we introduced a novel, and more accurate spin estimation approach.
    Finally, we show how combining the output of an event-based camera and a Spiking Neural Network (SNN) can be used for accurate ball detection.
\end{abstract}

\keywords{CoRL, RoboLetics, Table tennis robot}


\section{Introduction}

Table tennis is a fast-paced and exhilarating sport that demands agility, precision, and lightning-fast reflexes.
It is a sport enjoyed by millions of enthusiasts worldwide, ranging from casual players to professional athletes.
In recent years, the fusion of technology and sports has led to the development of various training aids and innovations aimed at enhancing the skills of players and fostering their competitive edge.
Among these technological advancements, table tennis robots have also emerged.
While not yet able to compete with professional players, table tennis robots are an interesting research environment to bring perception and control algorithms towards their limits.
Thus, it is not surprising, that more and more research groups use table tennis robots as a test bed for their algorithms~\citep{Tebbe2019gcpr}\citep{DAmbrosio2023rss}\citep{GomezGonzalez2019robotics}\citep{Ding2022iros}.

\section{Related Work}\label{sec:related_work}

Ever since Billingsley initiated a robot table tennis competition in 1983~\citep{Billingsley1983robot}, robotic table tennis has been a popular tool for research in computer vision and robot control.
Various types of manipulators have been used.
\citet{Huang2011iros} used a 5-DOF robot with three linear axes plus a pan-tilt unit.
\citet{Xiong2012ijoars} developed two human-like robots Wu \& Kong.
Both robots have 30 DOF in total with two 7-DOF arms, two 6-DOF legs, and 4-DOF for head and hip.
Omron frequently showcases its Delta robot at trade shows, with a table tennis racket attached after two additional swivel joints.
\citet{Buchler2022tor} have designed a completely new pneumatic robot arm able to attain very high-end-effector speeds.
Particularly popular are industrial 6 or 7-axis articulated arm robots in which all joints are rotational and which are relatively similar to the human arm~\citep{Mulling2013ijors}\citep{Lin2020sensors}\citep{Gao2020iros}.
Our system also employs this type of robot, the Agilus KR6 R900 sixx made by KUKA~\citep{Tebbe2021Tuebingen}.

For the perception of the table tennis ball, either conventional image processing techniques~\citep{Li2012icarcv} or Convolutional Neural Network (CNN) based approaches~\citep{GomezGonzalez2019robotics}\citep{Ding2022iros} are used.
While using a CNN for the ball detection results in a slightly higher accuracy~\citep{GomezGonzalez2019robotics}\cite{Tebbe2021Tuebingen} these approaches are more cumbersome to integrate and debug compared to their conventional counterparts~\cite{Tebbe2021Tuebingen}.
Therefore, we use a conventional image processing approach to detect the balls, introduced in~\citet{Tebbe2019gcpr} and later improved in~\citet{Tebbe2021Tuebingen}.
%

Since the spin of a table tennis ball is crucial in table tennis, we also aim to detect the spin with our perception system.
There are multiple ways to estimate the spin of a table tennis ball.
Model-based approaches use a physical model of the ball and estimate the spin by using the effect of the magnus force~\citep{Tebbe2020icra}.
As further research has shown, estimating the spin with a visual perception system leads to better results.
In most cases, the logo is used to determine the rotation of the ball and thus the spin~\citep{Tebbe2020icra}.

\section{Table Tennis Robot System}\label{sec:system_overview}

Our research is based on the table tennis robot system presented in~\citet{Tebbe2019gcpr}.
Which was further improved in~\citet{Tebbe2020icra} and~\citet{Tebbe2021Tuebingen}.
This system consists of four PointGrey Chameleon3 frame-based cameras (140 fps, 1280x1024 pixels), one PointGrey Grasshoper3 frame-based camera (350 fps, 1920x1200 pixels) and a 6-DOF KUKA Agilus robot.

We added accurate and reliable spin estimation using the high framerate camera, attached to the ceiling.
The algorithm used is described in \cref{subsec:spin_detection}.
We also extended the setup with two Prophesee EVK4 event-based cameras (1280x720 pixels), as event-cameras offer a lot of potential with the lower latency.
%
%
In \cref{subsec:calibration} we explain how we calibrated our setup to be able to use both types of camera simulteneously.
Ball positions estimated with frame cameras can then be used for learning-based method for event cameras as ground truth. 
This use of event cameras for table tennis is studied in \cref{subsec:snn_detection} where we describe a Spiking Neural Network (SNN) approach, using event-based data for ball detection.


\subsection{Spin Detection}\label{subsec:spin_detection}
\begin{wrapfigure}{r}{0.2\textwidth}
  \includegraphics[width=\linewidth]{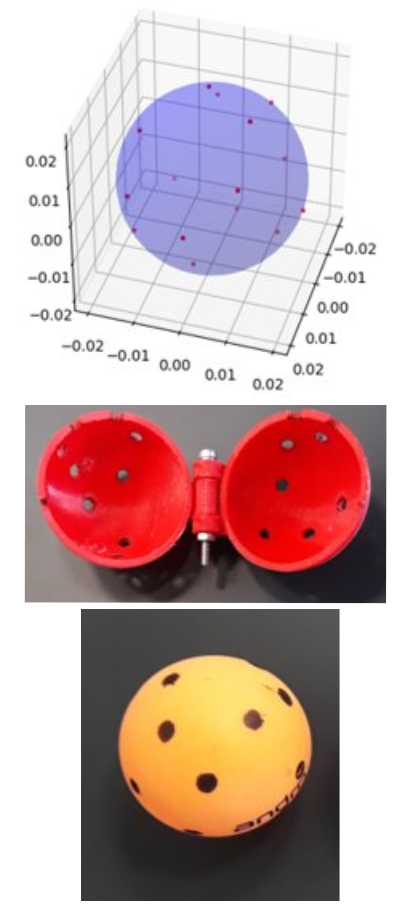}
  \caption{Table tennis balls with dot pattern and corresponding 3D printed stencil.}
  \label{fig:3d_print}
\end{wrapfigure}
For table tennis, spin estimation is crucial.
%
%
Several methods have been developed to solve this problem.
We focus on direct spin observation methods from ball images.
They can be separated into two categories: pattern-based \cite{Tamaki2004fcv} \cite{Theobalt2004siggraph}\cite{Furuno2009iccas}\cite{Szep2011cvww}\cite{Tamaki2012icassp} or logo-based \cite{Glover2014icra}\cite{Zhang2015tim}\cite{Tebbe2020icra}.
The logo-based methods can give out accurate spin estimations, but they fail when the logo is not visible, which makes them unreliable.
Previous pattern-based methods relied on registration, which requires high frame rate and high resolution cameras.
To overcome these problems, we implemented SpinDOE\cite{Gossard2023iros}. 
SpinDOE uses a dot pattern drawn onto the ball with a 3D printed stencil for accurate and repeatable dot positions. 

\begin{figure}[!htbp]
  \centering
  \includegraphics[width=0.6\linewidth]{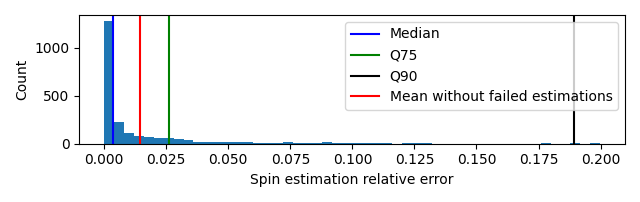}
  \caption{Spin estimation benchmark (failed estimation are those where the relative)}
  \label{fig:spin_benchmark}
\end{figure}

With our method, we can estimate spins up to 175 rps with a 350 fps camera and a 60x60 ball image resolution.
It provides great accuracy as it can be seen in \cref{fig:spin_benchmark}.
Failure cases come from angle unwrapping errors for extremely high or low spins.

The estimated spin can be used to adapt the racket's orientation for the returning stroke.
Our method also enabled us to empirically check the standard assumption in table tennis robotics that the spin is constant while the ball is airborne.
The spin dampening coefficient was estimated to be $k = 0.091 \pm 0.03$ with $\omega(t) =\omega(t_0) \exp(- k t )$.

\subsection{Multi-modal calibration}\label{subsec:calibration}

\begin{figure}[!htbp]
  \centering
  \includegraphics[width=0.7\linewidth]{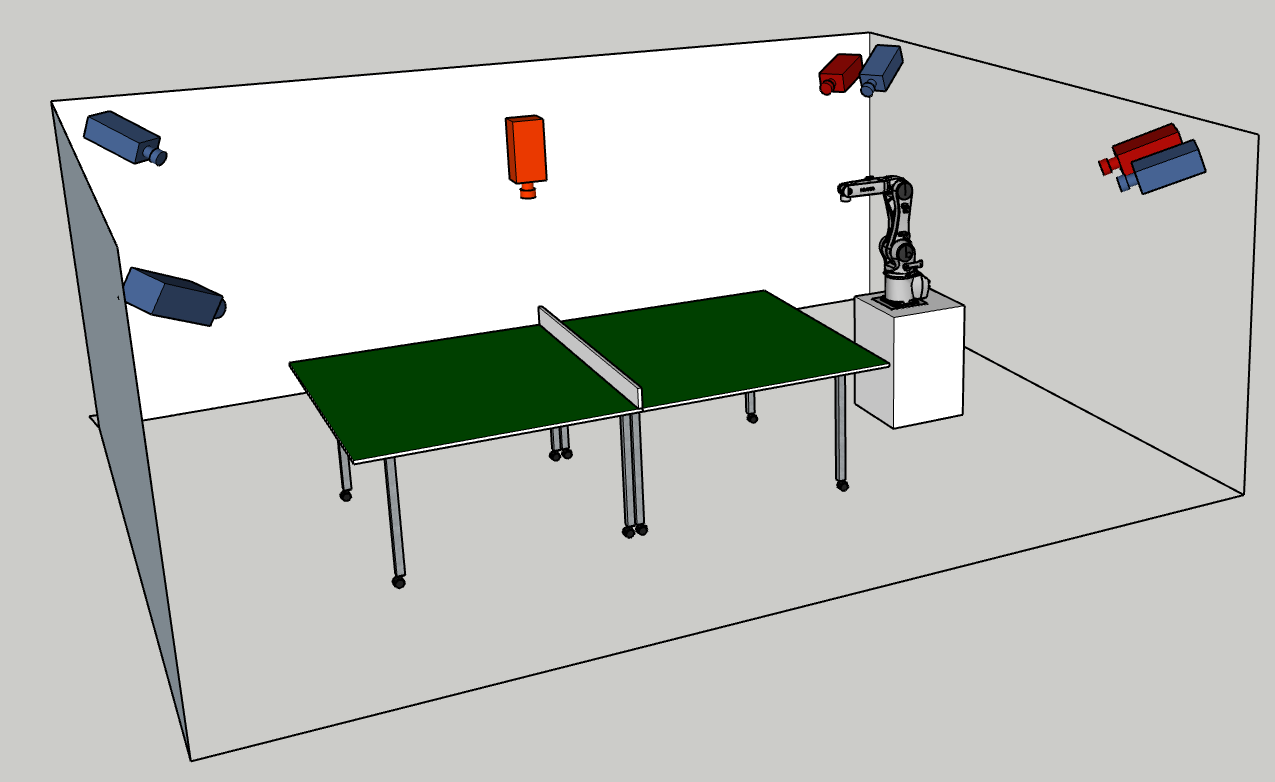}
  \caption{
    Our camera setup consisting of {\color{blue}four frame-based cameras (in blue)}, {\color{orange}one high frame-rate frame-based cameras for ball spin estimation (in orange)} and {\color{red}two event-based cameras (in red)} with baselines of $3$m to $5$m.
    Schematic is up to scale.
  }
  \label{fig:camera_setup}
\end{figure}

In~\citet{Gossard2023arxiv}, we introduced \textbf{eWand}, a wand-based calibration approach for frame-based and event-based cameras supporting wider baselines and cameras facing different fields of view.
%
Our setup is described in \cref{fig:camera_setup}.
Camera calibration for conventional cameras is a well-studied subject, with literature published over multiple decades~\cite{Clarke1998tpr}\cite{ZhengyouZhang1999iccv}\cite{Rehder2016icra}.
However, not many of these approaches are applicable for our setup.
Also, since event-based cameras only report illumination changes, such approaches cannot directly be applied to event-based cameras.
To overcome these limitations, we propose \textbf{eWand}~\citep{Gossard2023arxiv}, a new method that uses blinking LEDs inside opaque spheres instead of a printed or displayed pattern.
Our method provides a faster, easier-to-use extrinsic calibration approach that maintains high accuracy, listed in \cref{tab:reprojection_errors}, for both event- and frame-based cameras.
\begin{table*}[!ht]
    \centering
    \resizebox{\textwidth}{!}{
	\begin{tabular}{l|l|l|l|l|l|l}
		\textbf{Camera} & \textbf{BA eWand (our)} & \textbf{kalibr~\cite{Rehder2016icra} (circleboard)} & \textbf{kalibr~\cite{Rehder2016icra} (checkerboard)} & \textbf{BA circleboard} & \textbf{BA checkerboard} \\
		\toprule
        frame\_0 & $0.325 \pm 0.257$ & $0.274 \pm 0.060$ & $0.390 \pm 0.325$ & $0.218 \pm 0.219$ & $0.225 \pm 0.204$ \\ 
        frame\_1 & $0.320 \pm 0.271$ & $0.311 \pm 0.070$ & $0.398 \pm 0.335$ & $0.201 \pm 0.178$ & $0.201 \pm 0.185$ \\
        frame\_2 & $0.401 \pm 0.361$ & $0.168 \pm 0.072$ & $0.320 \pm 0.320$ & $0.178 \pm 0.178$ & $0.213 \pm 0.193$ \\
        frame\_3 & $0.396 \pm 0.389$ & $0.159 \pm 0.062$ & $0.301 \pm 0.243$ & $0.199 \pm 0.188$ & $0.180 \pm 0.165$ \\
        event\_0 & $0.603 \pm 0.448$ & $0.269 \pm 0.068$ & $0.513 \pm 0.396$ & $0.398 \pm 0.365$ & $0.459 \pm 0.363$ \\ 
        event\_1 & $0.550 \pm 0.431$ & $0.295 \pm 0.070$ & $0.515 \pm 0.405$ & $0.469 \pm 0.447$ & $0.402 \pm 0.370$ \\
		\hline
	\end{tabular}
    }
    \caption{Reprojection error (MAE) in pixels (mean and std.) for each calibration approach and camera, after the calibration.
             "BA" represents a bundle adjustment approach using OpenCV and Ceres.}
    \label{tab:reprojection_errors}
\end{table*}

\subsection{SNN for event-based ball detection}\label{subsec:snn_detection}

Real-time table-tennis ball tracking, crucial for enabling a robotic arm to rally the ball back successfully, demands both fast and accurate ball detection.
Previous approaches have employed frame-based cameras with CNNs or traditional computer vision methods.
We enter a different avenue and combine event-based cameras and a Spiking Neural Network (SNN) for ball detection.

SNNs mimic the spiking behavior of biological neurons, offering a biologically inspired approach to neural network computation.
Unlike traditional artificial neurons that produce real-valued outputs, spiking neurons accumulate input in a membrane potential until a threshold is reached, triggering a spike.
This binary output format of SNNs is a perfect fit for event-based cameras and enables energy-efficient computations.

\begin{figure}[!ht]
  \centering
  \includegraphics[width=0.45\linewidth]{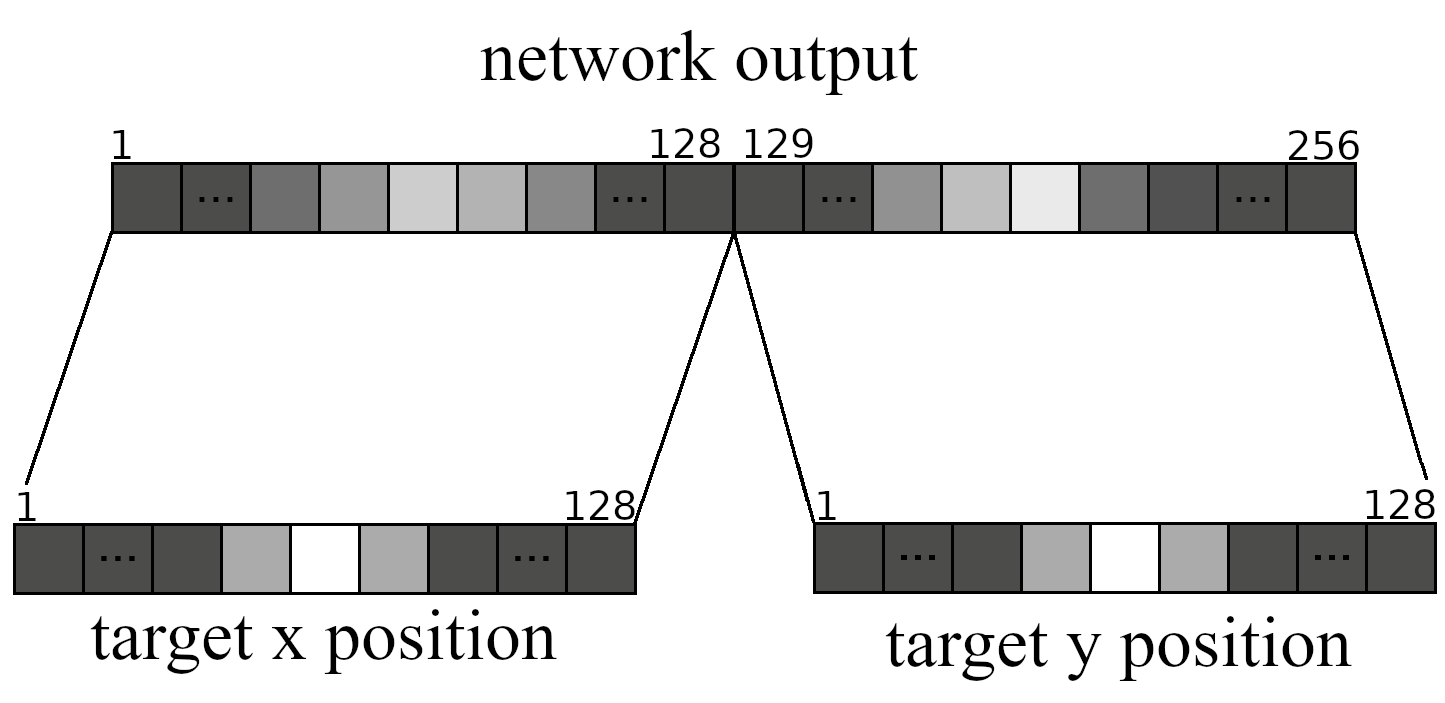}
  \caption{
    The network has 256 output neurons, which are split into two populations.
    Each of the neurons in the first population represents a $x$ position, each of the neurons in the second a $y$ position.
    The target is also split into a $x$ and $y$ target, each setting one as the target activity for the correct neuron, $0.5$ for the two adjacent neurons and zero for all others.
    Values are represented using brightness, with larger values being brighter.
  }
  \label{fig:snn_output}
\end{figure}

In our approach, we treat $x$ and $y$ positions as two independent classification tasks.
In this case, the ball's $x$ position can be one of 128 classes, as can the $y$ position.
A visualization of the network's output is provided in \cref{fig:snn_output}.

The network consists of four layers, with the first two being convolutional layers and the last two being linear layers.
The heaviside step function is used as activation function.
%
%

While the standard approach would be to use the cross entropy loss function to train a classification network, the mean squared error (MSE) proved more suitable to this problem.
The cause of this is not poor accuracy when training with cross entropy loss, but the tendency of the cross entropy loss to produce networks with very large activations.
Large activations result in more spikes, causing more synaptic operations.
%

In \cref{tab:snn_results} we list the results of the SNN based ball detection.
\begin{table}[!ht]
    \centering
    \begin{tabular}{ccc}
        \centering
        Time steps & Accuracy [pixels] & SynOps \\
        \hline
        8 & $1.00 \pm 0.04$ & $1.43 \cdot 10^5$ \\
        \hline
        16 & $0.92 \pm 0.03$ & $2.97 \cdot 10^5$ \\
        \hline
        32 & $\mathbf{0.89} \pm \mathbf{0.03}$ & $6.06 \cdot 10^5$ \\
        \hline
    \end{tabular}
    \caption{
    Average Euclidean distance in pixels and standard deviation across 10 networks.
    %
    %
    Also shown are the average number of synaptic operations (SynOps) per forward pass.
    Numbers in bold mark the best performing method.
    }
    \label{tab:snn_results}
\end{table}
As a comparison, in~\citet{Tebbe2019gcpr}, we achieved an accuracy of $1.53$ pixels with a classical computer vision approach on frames from a frame-based camera.




	


\section{Conclusion}
\label{sec:conclusion}

In this paper, we started with an introduction of our table tennis robot system, originally presented in~\citep{Tebbe2019gcpr}.
We then highlighted our recent advancements, ranging from a novel calibration approach to calibrate our multimodal perception system, over a more accurate spin estimation approach, to a Spiking Neural Network (SNN) approach for ball detection with event-based cameras.
We hope that this work will foster the discussion, how table tennis robot systems can be further improved and maybe compete against professional players at one point.


\clearpage
\acknowledgments{
We would like to thank Kin Man Lee and Zulfiqar Zaidi for approaching us and their encouragement to submit our recent developments. 
We also appreciate the valuable feedback from Timon Höfer, Thomas Ziegler, and Philipp Riedel.
Special thanks to Sony AI for funding this project.
}


\bibliography{bibliography}  

\begin{thebibliography}{26}
\providecommand{\natexlab}[1]{#1}
\providecommand{\url}[1]{\texttt{#1}}
\expandafter\ifx\csname urlstyle\endcsname\relax
  \providecommand{\doi}[1]{doi: #1}\else
  \providecommand{\doi}{doi: \begingroup \urlstyle{rm}\Url}\fi

\bibitem[Tebbe et~al.(2019)Tebbe, Gao, Sastre-Rienietz, and
  Zell]{Tebbe2019gcpr}
J.~Tebbe, Y.~Gao, M.~Sastre-Rienietz, and A.~Zell.
\newblock A table tennis robot system using an industrial {KUKA} robot arm.
\newblock In \emph{Lecture Notes in Computer Science}, pages 33--45. Springer
  International Publishing, 2019.
\newblock \doi{10.1007/978-3-030-12939-2_3}.
\newblock URL \url{https://doi.org/10.1007/978-3-030-12939-2_3}.

\bibitem[D{\textquotesingle}Ambrosio et~al.(2023)D{\textquotesingle}Ambrosio,
  Jaitly, Sindhwani, Oslund, Xu, Lazic, Shankar, Ding, Abelian, Coumans,
  Kouretas, Nguyen, Boyd, Iscen, Mahjourian, Vanhoucke, Bewley, Kuang, Ahn,
  Jain, Kataoka, Cortes, Sermanet, Lynch, Sanketi, Choromanski, Gao,
  Kangaspunta, Reymann, Vesom, Moore, Singh, Abeyruwan, and
  Graesser]{DAmbrosio2023rss}
D.~D{\textquotesingle}Ambrosio, N.~Jaitly, V.~Sindhwani, K.~Oslund, P.~Xu,
  N.~Lazic, A.~Shankar, T.~Ding, J.~Abelian, E.~Coumans, G.~Kouretas,
  T.~Nguyen, J.~Boyd, A.~Iscen, R.~Mahjourian, V.~Vanhoucke, A.~Bewley,
  Y.~Kuang, M.~Ahn, D.~Jain, S.~Kataoka, O.~Cortes, P.~Sermanet, C.~Lynch,
  P.~Sanketi, K.~Choromanski, W.~Gao, J.~Kangaspunta, K.~Reymann, G.~Vesom,
  S.~Moore, A.~Singh, S.~Abeyruwan, and L.~Graesser.
\newblock Robotic table tennis: A case study into a high speed learning system.
\newblock In \emph{Robotics: Science and Systems {XIX}}. Robotics: Science and
  Systems Foundation, July 2023.
\newblock \doi{10.15607/rss.2023.xix.006}.
\newblock URL \url{https://doi.org/10.15607/rss.2023.xix.006}.

\bibitem[Gomez-Gonzalez et~al.(2019)Gomez-Gonzalez, Nemmour, Sch\"{o}lkopf, and
  Peters]{GomezGonzalez2019robotics}
S.~Gomez-Gonzalez, Y.~Nemmour, B.~Sch\"{o}lkopf, and J.~Peters.
\newblock Reliable real-time ball tracking for robot table tennis.
\newblock \emph{Robotics}, 8\penalty0 (4):\penalty0 90, Oct. 2019.
\newblock \doi{10.3390/robotics8040090}.
\newblock URL \url{https://doi.org/10.3390/robotics8040090}.

\bibitem[Ding et~al.(2022)Ding, Graesser, Abeyruwan,
  D{\textquotesingle}Ambrosio, Shankar, Sermanet, Sanketi, and
  Lynch]{Ding2022iros}
T.~Ding, L.~Graesser, S.~Abeyruwan, D.~B. D{\textquotesingle}Ambrosio,
  A.~Shankar, P.~Sermanet, P.~R. Sanketi, and C.~Lynch.
\newblock Learning high speed precision table tennis on a physical robot.
\newblock In \emph{2022 {IEEE}/{RSJ} International Conference on Intelligent
  Robots and Systems ({IROS})}. {IEEE}, Oct. 2022.
\newblock \doi{10.1109/iros47612.2022.9982205}.
\newblock URL \url{https://doi.org/10.1109/iros47612.2022.9982205}.

\bibitem[Billingsley(1983)]{Billingsley1983robot}
J.~Billingsley.
\newblock Robot ping pong.
\newblock \emph{Practical Computing}, 6\penalty0 (5), 1983.

\bibitem[Huang et~al.(2011)Huang, Xu, Tan, and Su]{Huang2011iros}
Y.~Huang, D.~Xu, M.~Tan, and H.~Su.
\newblock Trajectory prediction of spinning ball for ping-pong player robot.
\newblock In \emph{2011 {IEEE}/{RSJ} International Conference on Intelligent
  Robots and Systems}. {IEEE}, Sept. 2011.
\newblock \doi{10.1109/iros.2011.6095044}.
\newblock URL \url{https://doi.org/10.1109/iros.2011.6095044}.

\bibitem[Xiong et~al.(2012)Xiong, Sun, Zhu, Wu, and Chu]{Xiong2012ijoars}
R.~Xiong, Y.~Sun, Q.~Zhu, J.~Wu, and J.~Chu.
\newblock Impedance control and its effects on a humanoid robot playing table
  tennis.
\newblock \emph{International Journal of Advanced Robotic Systems}, 9\penalty0
  (5):\penalty0 178, Nov. 2012.
\newblock \doi{10.5772/51924}.
\newblock URL \url{https://doi.org/10.5772/51924}.

\bibitem[Buchler et~al.(2022)Buchler, Guist, Calandra, Berenz, Scholkopf, and
  Peters]{Buchler2022tor}
D.~Buchler, S.~Guist, R.~Calandra, V.~Berenz, B.~Scholkopf, and J.~Peters.
\newblock Learning to play table tennis from scratch using muscular robots.
\newblock \emph{{IEEE} Transactions on Robotics}, 38\penalty0 (6):\penalty0
  3850--3860, Dec. 2022.
\newblock \doi{10.1109/tro.2022.3176207}.
\newblock URL \url{https://doi.org/10.1109/tro.2022.3176207}.

\bibitem[M\"{u}lling et~al.(2013)M\"{u}lling, Kober, Kroemer, and
  Peters]{Mulling2013ijors}
K.~M\"{u}lling, J.~Kober, O.~Kroemer, and J.~Peters.
\newblock Learning to select and generalize striking movements in robot table
  tennis.
\newblock \emph{The International Journal of Robotics Research}, 32\penalty0
  (3):\penalty0 263--279, Jan. 2013.
\newblock \doi{10.1177/0278364912472380}.
\newblock URL \url{https://doi.org/10.1177/0278364912472380}.

\bibitem[Lin et~al.(2020)Lin, Yu, and Huang]{Lin2020sensors}
H.-I. Lin, Z.~Yu, and Y.-C. Huang.
\newblock Ball tracking and trajectory prediction for table-tennis robots.
\newblock \emph{Sensors}, 20\penalty0 (2):\penalty0 333, Jan. 2020.
\newblock \doi{10.3390/s20020333}.
\newblock URL \url{https://doi.org/10.3390/s20020333}.

\bibitem[Gao et~al.(2020)Gao, Graesser, Choromanski, Song, Lazic, Sanketi,
  Sindhwani, and Jaitly]{Gao2020iros}
W.~Gao, L.~Graesser, K.~Choromanski, X.~Song, N.~Lazic, P.~Sanketi,
  V.~Sindhwani, and N.~Jaitly.
\newblock Robotic table tennis with model-free reinforcement learning.
\newblock In \emph{2020 {IEEE}/{RSJ} International Conference on Intelligent
  Robots and Systems ({IROS})}. {IEEE}, Oct. 2020.
\newblock \doi{10.1109/iros45743.2020.9341191}.
\newblock URL \url{https://doi.org/10.1109/iros45743.2020.9341191}.

\bibitem[Tebbe(2021)]{Tebbe2021Tuebingen}
J.~Tebbe.
\newblock \emph{Adaptive robot systems in highly dynamic environments: A table
  tennis robot}.
\newblock PhD thesis, University of Tübingen, 2021.

\bibitem[Li et~al.(2012)Li, Wu, Lou, Kuhnlenz, and Ravn]{Li2012icarcv}
H.~Li, H.~Wu, L.~Lou, K.~Kuhnlenz, and O.~Ravn.
\newblock Ping-pong robotics with high-speed vision system.
\newblock In \emph{2012 12th International Conference on Control Automation
  Robotics \& Vision (ICARCV)}. {IEEE}, Dec. 2012.
\newblock \doi{10.1109/icarcv.2012.6485142}.
\newblock URL \url{https://doi.org/10.1109/icarcv.2012.6485142}.

\bibitem[Tebbe et~al.(2020)Tebbe, Klamt, Gao, and Zell]{Tebbe2020icra}
J.~Tebbe, L.~Klamt, Y.~Gao, and A.~Zell.
\newblock Spin detection in robotic table tennis.
\newblock In \emph{2020 {IEEE} International Conference on Robotics and
  Automation ({ICRA})}. {IEEE}, May 2020.
\newblock \doi{10.1109/icra40945.2020.9196536}.
\newblock URL \url{https://doi.org/10.1109/icra40945.2020.9196536}.

\bibitem[Tamaki et~al.(2004)Tamaki, Sugino, and Yamamoto]{Tamaki2004fcv}
T.~Tamaki, T.~Sugino, and M.~Yamamoto.
\newblock Measuring ball spin by image registration.
\newblock \emph{Proc. 10th Frontiers of Computer Vision}, pages 269--274, 2004.

\bibitem[Theobalt et~al.(2004)Theobalt, Albrecht, Haber, Magnor, and
  Seidel]{Theobalt2004siggraph}
C.~Theobalt, I.~Albrecht, J.~Haber, M.~Magnor, and H.-P. Seidel.
\newblock Pitching a baseball.
\newblock \emph{{ACM} Transactions on Graphics}, 23\penalty0 (3):\penalty0
  540--547, Aug. 2004.
\newblock \doi{10.1145/1015706.1015758}.
\newblock URL \url{https://doi.org/10.1145/1015706.1015758}.

\bibitem[Furuno et~al.(2009)Furuno, Kobayashi, Okubo, and
  Kurihara]{Furuno2009iccas}
S.~Furuno, K.~Kobayashi, T.~Okubo, and Y.~Kurihara.
\newblock A study on spin-rate measurement using a uniquely marked moving ball.
\newblock In \emph{2009 ICCAS-SICE}, pages 3439--3442, 2009.

\bibitem[Szep(2011)]{Szep2011cvww}
A.~Szep.
\newblock Measuring ball spin in monocular video.
\newblock In \emph{Proc. 16th Comput. Vis. Winter Workshop}, pages 83--89.
  Citeseer, 2011.

\bibitem[Tamaki et~al.(2012)Tamaki, Wang, Raytchev, Kaneda, and
  Ushiyama]{Tamaki2012icassp}
T.~Tamaki, H.~Wang, B.~Raytchev, K.~Kaneda, and Y.~Ushiyama.
\newblock Estimating the spin of a table tennis ball using inverse
  compositional image alignment.
\newblock In \emph{2012 {IEEE} International Conference on Acoustics, Speech
  and Signal Processing ({ICASSP})}. {IEEE}, Mar. 2012.
\newblock \doi{10.1109/icassp.2012.6288166}.
\newblock URL \url{https://doi.org/10.1109/icassp.2012.6288166}.

\bibitem[Glover and Kaelbling(2014)]{Glover2014icra}
J.~Glover and L.~P. Kaelbling.
\newblock Tracking the spin on a ping pong ball with the quaternion bingham
  filter.
\newblock In \emph{2014 {IEEE} International Conference on Robotics and
  Automation ({ICRA})}. {IEEE}, May 2014.
\newblock \doi{10.1109/icra.2014.6907460}.
\newblock URL \url{https://doi.org/10.1109/icra.2014.6907460}.

\bibitem[Zhang et~al.(2015)Zhang, Xiong, Zhao, and Wang]{Zhang2015tim}
Y.~Zhang, R.~Xiong, Y.~Zhao, and J.~Wang.
\newblock Real-time spin estimation of ping-pong ball using its natural brand.
\newblock \emph{{IEEE} Transactions on Instrumentation and Measurement},
  64\penalty0 (8):\penalty0 2280--2290, Aug. 2015.
\newblock \doi{10.1109/tim.2014.2385173}.
\newblock URL \url{https://doi.org/10.1109/tim.2014.2385173}.

\bibitem[Gossard et~al.(2023{\natexlab{a}})Gossard, Tebbe, Ziegler, and
  Zell]{Gossard2023iros}
T.~Gossard, J.~Tebbe, A.~Ziegler, and A.~Zell.
\newblock Spindoe: A ball spin estimation method for table tennis robot.
\newblock In \emph{IEEE/RSJ Int. Conf. Intell. Robot. Syst. (IROS)}. IEEE,
  2023{\natexlab{a}}.
\newblock \doi{10.48550/ARXIV.2303.03879}.
\newblock URL \url{https://arxiv.org/abs/2303.03879}.

\bibitem[Gossard et~al.(2023{\natexlab{b}})Gossard, Ziegler, Kolmar, Tebbe, and
  Zell]{Gossard2023arxiv}
T.~Gossard, A.~Ziegler, L.~Kolmar, J.~Tebbe, and A.~Zell.
\newblock ewand: A calibration framework for wide baseline frame-based and
  event-based camera systems, 2023{\natexlab{b}}.

\bibitem[Clarke and Fryer(1998)]{Clarke1998tpr}
T.~A. Clarke and J.~G. Fryer.
\newblock The development of camera calibration methods and models.
\newblock \emph{The Photogrammetric Record}, 16\penalty0 (91):\penalty0 51--66,
  Apr. 1998.
\newblock \doi{10.1111/0031-868x.00113}.
\newblock URL \url{https://doi.org/10.1111/0031-868x.00113}.

\bibitem[Zhang(1999)]{ZhengyouZhang1999iccv}
Z.~Zhang.
\newblock Flexible camera calibration by viewing a plane from unknown
  orientations.
\newblock In \emph{Proceedings of the Seventh {IEEE} International Conference
  on Computer Vision}. {IEEE}, 1999.
\newblock \doi{10.1109/iccv.1999.791289}.
\newblock URL \url{https://doi.org/10.1109/iccv.1999.791289}.

\bibitem[Rehder et~al.(2016)Rehder, Nikolic, Schneider, Hinzmann, and
  Siegwart]{Rehder2016icra}
J.~Rehder, J.~Nikolic, T.~Schneider, T.~Hinzmann, and R.~Siegwart.
\newblock Extending kalibr: Calibrating the extrinsics of multiple {IMUs} and
  of individual axes.
\newblock In \emph{2016 {IEEE} International Conference on Robotics and
  Automation ({ICRA})}. {IEEE}, May 2016.
\newblock \doi{10.1109/icra.2016.7487628}.
\newblock URL \url{https://doi.org/10.1109/icra.2016.7487628}.

\end{thebibliography}

\end{document}